\documentclass[10pt,twocolumn,letterpaper]{article}

\usepackage{cvpr}
\usepackage{times}
\usepackage{epsfig}
\usepackage{graphicx}
\usepackage{amsmath}
\usepackage{amssymb}

\usepackage{algorithm}  
\usepackage{algpseudocode}  
\usepackage{amsmath}  
\usepackage{color}
\usepackage{multirow}
\usepackage[breaklinks=true,bookmarks=false]{hyperref}

\usepackage[T1]{fontenc}
\usepackage[utf8]{inputenc}
\usepackage{authblk}

\cvprfinalcopy 

\def\cvprPaperID{07} 

\ifcvprfinal\fi
\begin{document}

\title{PipeNet: Selective Modal Pipeline of Fusion Network \\for Multi-Modal Face Anti-Spoofing}

\author[1]{Qing Yang}
\author[2]{Xia Zhu}
\author[2]{Jong-Kae Fwu}
\author[1]{Yun Ye}
\author[1]{Ganmei You}
\author[1]{Yuan Zhu}
\affil[1]{Intel, China}
\affil[1]{\{qing.y.yang, yun.ye, ganmei.you, yuan.y.zhu\}@intel.com}
\affil[2]{Intel, USA}
\affil[2]{\{xia.zhu, jong-kae.fwu\}@intel.com}
\maketitle

\begin{abstract}
Face anti-spoofing has become an increasingly important and critical security feature for authentication systems, due to rampant and easily launchable presentation attacks. 
Addressing the shortage of multi-modal face dataset, CASIA recently released the largest up-to-date CASIA-SURF Cross-ethnicity Face Anti-spoofing(CeFA) dataset, covering 3 ethnicities, 3 modalities, 1607 subjects, and 2D plus 3D attack types in four protocols, and focusing on the challenge of improving the generalization capability of face anti-spoofing in cross-ethnicity and multi-modal continuous data.
In this paper, we propose a novel pipeline-based multi-stream CNN architecture called PipeNet for multi-modal face anti-spoofing. Unlike previous works, Selective Modal Pipeline (SMP) is designed to enable a customized pipeline for each data modality to take full advantage of multi-modal data. Limited Frame Vote (LFV) is designed to ensure stable and accurate prediction for video classification.   
The proposed method wins the third place in the final ranking of Chalearn Multi-modal Cross-ethnicity Face Anti-spoofing Recognition Challenge@CVPR2020. Our final submission achieves the Average Classification Error Rate (ACER) of $2.21\pm 1.26$ on the test set.
\end{abstract}

\section{Introduction}
 
Face recognition technology has been widely used in a variety of applications in our daily life, such as mobile face payment, entrance authentication and office check-in machines, etc. Unfortunately, human face's easy accessibility brings not only convenience but also presentation attacks. These facial recognition systems are vulnerable to various types of presentation attacks, including printed photograph, digital video replay, 3D print mask, and silica gel face \etal. Therefore, how to detect these various means of presentation attacks has become an increasingly critical and challenging task in all face recognition and authentication systems. 

Various face anti-spoofing methods have been proposed over the past decade. Traditional methods use the combination of handcrafted features~\cite{slrb,fbn6,fbn17,fbn19,fbn29,fbn21,fbn13,ftn1,ftn3,ftn4} such as Sparse Low Rank Bilinear (SLRB), Local Binary Patterns (LBP), Histogram of Oriented Gradients (HOG), Difference of Gaussian (DoG), Scale Invariant Feature Transform (SIFT), Speeded-Up Robust Feature (SURF) and traditional classifiers including Support Vector Machines (SVM) and Linear Discriminant Analysis (LDA) to solve the classification problem of real and fake face. However, due to poor generalization capability, it is difficult to apply these methods in changing illumination environments.  

Recently, deep convolutional neural networks (CNNs) have been widely used for various face related tasks such as face detection, recognition, identification and landmark detection \etal. Researchers have also introduced CNNs into face anti-spoofing and liveness detection area. It is proved that CNNs~\cite{ftn6,ftn7,ftn8,cnn1} can effectively extract richer deep facial image semantic features than traditional methods for binary classification of face anti-spoofing tasks. However, the single modal data input and supervised-learning mechanism makes it difficult to perform well in cross-dataset tests. 

\begin{figure}[htbp]
	\centering
	\includegraphics[height=3.5cm,width=6cm]{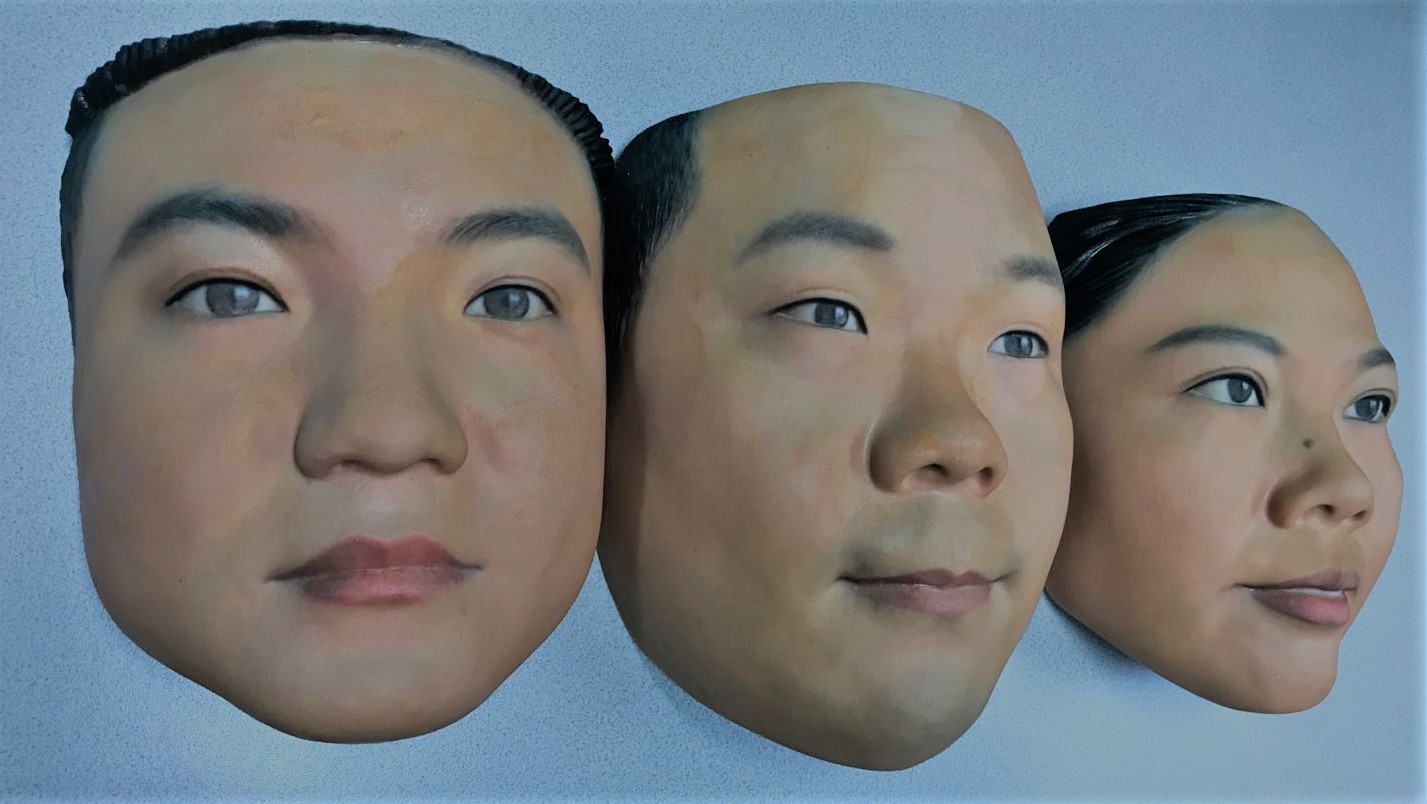}
	\caption{Low-cost high-precision 3D face masks.}
\label{fig:mask}
\end{figure}

RGB data can provide texture information, but it is insensitive to surface curvature and reflectivity of materials.
To improve the accuracy of face anti-spoofing systems (FAS), more modalities data such as depth~\cite{patch} and InfraRed (IR) are used as algorithm data input. Depth camera like Intel RealSense D400 series\footnote{https://www.intelrealsense.com/} can provide distance/depth information to make it easier to distinguish real face surfaces from face shapes in photograph and video replays on electronic displays. IR camera is sensitive to material reflectivity of different presentation attack instruments (PAI). Especially, fusion methods~\cite{visionlab,facebagnet,feathernets} become main stream with the large-scale multi-modal dataset CASIA-SURF~\cite{surf} publicly released.

Unfortunately, with the rapid development of presentation attack methods, face anti-spoofing systems have to deal with more challenging situations. 
As shown in Fig~\ref{fig:mask}, it becomes easier and cheaper to produce 3D printed masks, while it's extremely difficult to distinguish them from real faces in both RGB and depth data modalities.  In addition, silica gel face is more lifelike and it looks like real face in all $3$ modalities (RGB, depth and IR). For most previous approaches, it's impossible to distinguish between these two presentation attacks and live face. 

To address this issue, we propose a novel pipeline-based fusion framework which can effectively and respectively extract richer information from face data in different modalities. 

In summary, the main contributions of this paper are summarized as follows:

(1) We present an effective pipeline-based framework PipeNet for multi-modal face anti-spoofing task;

(2) In contrast to previous unified designs, we propose a Selective Modal Pipeline (SMP) design for differentiating feature extraction among different modalities data;

(3) We propose a novel method - Limited Frame Vote (LFV) to get stable and reliable prediction for video classification.

The rest of the paper is organized as follows: Section~\ref{related} briefly reviews related works in multi-modal face anti-spoofing; Section~\ref{cefa} introduces CASIA-SURF CeFA dataset and baseline methods; Section~\ref{method} proposes our method PipeNet with SMP and LFV modules; Section~\ref{competition} shows competition basic rules; Section~\ref{experiment} shows the experiments and results analysis; Finally, we summarize our work and future research direction in Section~\ref{conclusion}.

\section{Related Work}
\label{related}

\noindent \textbf{Traditional Methods}:
The facial motions in a sequence of frames are firstly utilized as cues in face anti-spoofing task. For example, eyes blinking is a reliable evidence that the face is real~\cite{do30,do36}.

In contrast to requesting user's cooperation, some existing approaches treat face anti-spoofing task as a binary classification problem which can proceed with single frame. Various hand-crafted features have been explored and adopted in previous works, such as SLRB, LBP, HOG, DOG, SIFT, and SURF. Followed by traditional classifiers such as SVM, LDA and Random Forest. 

\begin{table*}[htbp]
\begin{center}
\centering
\begin{tabular}{|l|c|c|c|c|c|c|c|c|c|}
\hline
Subset & \multicolumn{3}{|c|}{Ethnicities} & Subjects & Modalities & PAIs & \# real videos & \# fake videos & \# all videos\\
\hline\hline
~&4@1 & 4@2 & 4@3&
\multicolumn{6}{|c|}{} \\
\hline
  Train & A & C & E & 1-200 & R\&D\&I & Replay & 600/600/600 & 600/600/600 & 1200/1200/1200\\
 \hline
 Valid & A & C & E & 201-300 & R\&D\&I & Replay & 300/300/300 & 300/300/300 & 600/600/600\\
 \hline
 Test & C\&E & A\&E & A\&C & 301-500 & R\&D\&I & Print & 1200/1200/1200 & 5400/5400/5400 & 6600/6600/6600\\
\hline
\end{tabular}
\end{center}
\caption{CASIA SURF CeFA dataset Protocol 4.}
\label{tab:cefa}
\end{table*}

\noindent \textbf{CNN-based Methods}:
After AlexNet's~\cite{NIPS2012_4824} great success, the strong representation capability of CNNs has been exploited in face anti-spoofing research~\cite{ftn6,ftn7,ftn8,data42}. CNN methods~\cite{cnn1} attempt to learn deep feature representations from face image data for binary classification. Two-stream CNNs~\cite{patch} are proposed to overcome the weakness of cross-dataset test performance based on both patch and depth streams. Li~\etal~\cite{ftn7} proposed a method to link partial features and Principle Component Analysis (PCA), instead of fully-connected layer, to reduce the dimensionality of features in order to avoid the over-fitting problem.

\noindent \textbf{Temporal-based Methods}:
Temporal information is also used to enhance CNN's representation capability. 
Long Short Term Memory (LSTM)~\cite{data,data41} can recurrently learn features to obtain context information, but the heavy computation overhead makes it difficult to deploy such method, especially on mobile devices. Remote photoplethysmography (rPPG)~\cite{ppg16,ppg17,ppg3d} measures a pulse signal of heart rate which only exists in live face, and can be used to 3D masks attack. However, weak robustness makes rPPG vulnerable to illumination changes of environment. More importantly, both methods take a long time for one testing shot, which is unpractical for deployment.

\noindent \textbf{Fusion-based Methods}:
Multi-Modal methods can take advantage of complementary information among different modalities data and provide more robust performance on face anti-spoofing task. However, there used to be only small face anti-spoofing dataset with limited subjects and samples available, which make it easy to fall into overfitting problem during CNN training.

In 2019, Zhang~\cite{surf} released CASIA-SURF, a large-scale multi-modal dataset for face anti-spoofing, which consists 1000 subjects with 21,000 videos and each sample with $3$ modalites data(\ie, RGB, depth and IR). 

With CASIA-SURF dataset, they hold the first round of Face Anti-spoofing Attack Detection Challenge@CVPR2019 ~\cite{2019w}. The top-3 solutions were as follows: 

VisionLab~\cite{visionlab} was the champion with ACER score of $0.0810\%$. They proposed a fusion network with multi-level feature aggregation module which can fully utilize the feature fusion from different modalities at both coarse and fine levels.
FaceBagNet~\cite{facebagnet} won the second place with ACER score of $0.0985\%$. They proposed a patch-based multi-stream fusion CNN architecture based on Bag-of-local-features. The patch-level images contribute to extract spoof-specific discriminative information and Model Feature Erasing module randomly erases one modal to prevent overfitting. 
FeatherNets~\cite{feathernets} was the third winner with ACER score of $0.1292\%$. They proposed a light-weighted network architecture with modified Global Average Pooling(GAP) named streaming module. Their fusion procedure is based on \textit{ensemble + cascade} structure to make best use of each modal data.

\begin{figure}[htbp]
	\centering
	\includegraphics[height=8.5 cm,width=8.5cm]{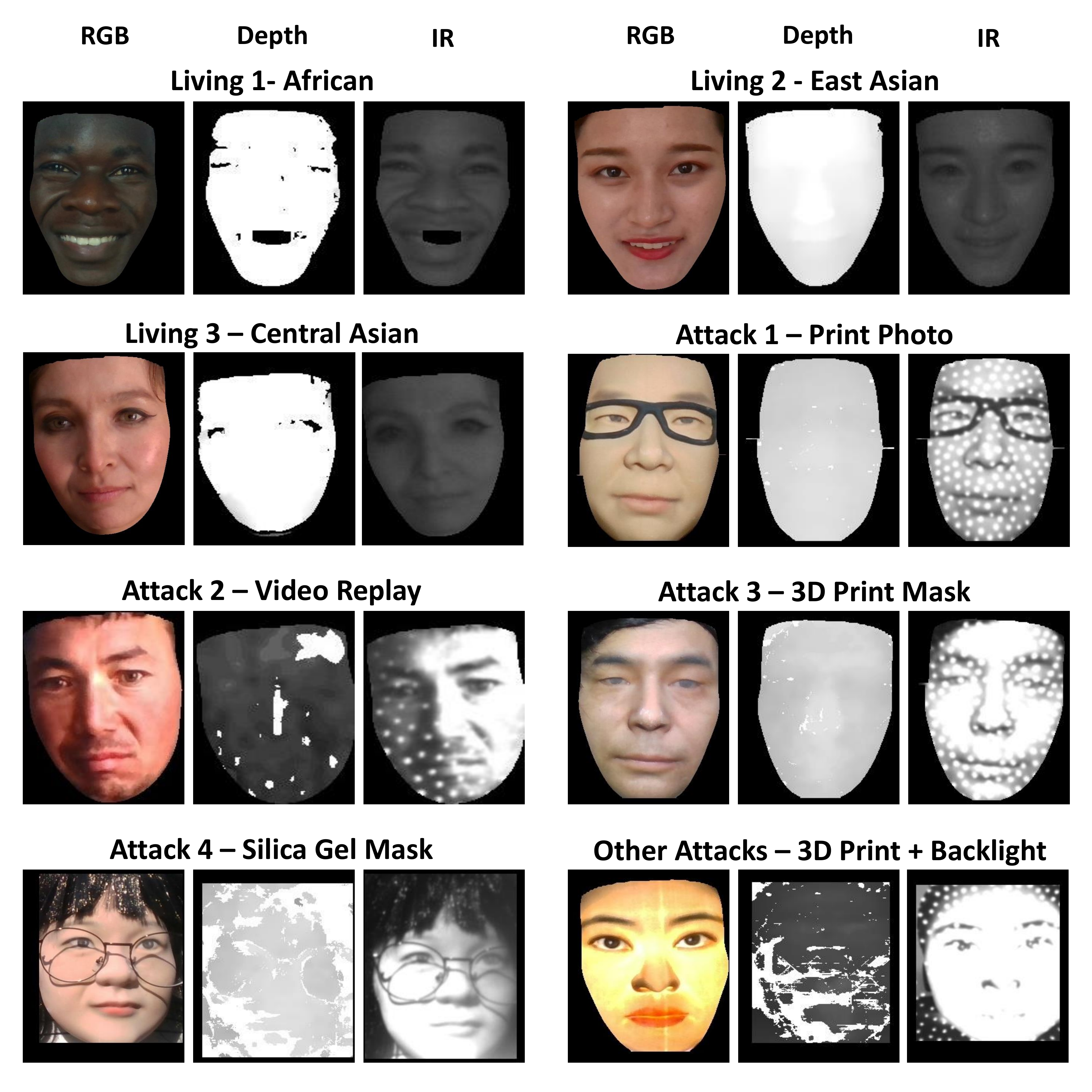}
	\caption{Examples of living and spoofing faces among different ethnicities and different attack types in three modalities (RGB, Depth, IR) from CASIA SURF-CeFA dataset~\cite{cefa}. }
\label{fig:cefa}
\end{figure}

\begin{figure*}
	\begin{center}
		\centering
		\includegraphics[height=8.5cm,width=18cm]{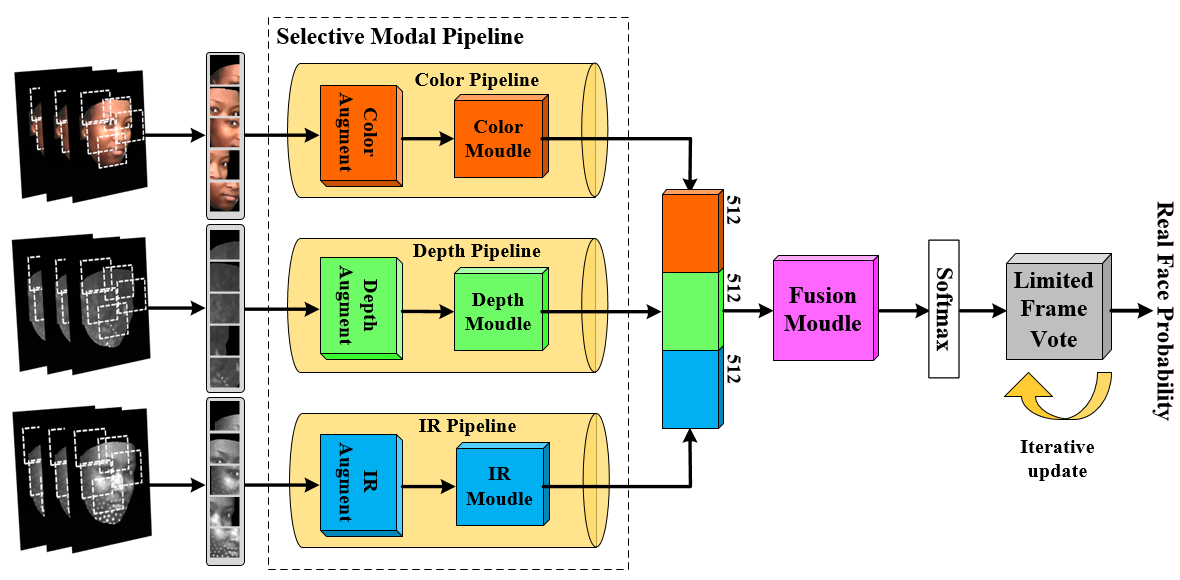}
	\end{center}
	\caption{The overall model architecture of PipeNet with Selective Modal Pipeline (SMP) module and Limited Frame Vote (LFV) module. Face patches of three modalities data are fed into corresponding RGB, Depth and IR pipelines in SMP module. The outputs of SMP are concatenated and sent into fusion module for further feature abstraction. LFV is applied to calculate the final output.}
\label{fig:arch}
\end{figure*}
\section{Dataset And Baseline}
\label{cefa}
\subsection{CASIA-SURF CeFA}
CASIA-SURF dataset only contains faces in one ethnicity. In order to provide a benchmark platform for improving the generalization capability in \textit{cross-ethnicity} anti-spoofing and the utilization of multi-modal continuous data, CASIA further released the largest up-to-date \textit{cross-ethnicity} face anti-spoofing dataset, CASIA-SURF CeFA~\cite{cefa} in 2020. CASIA-SURF CeFA covers $3$ ethnicities, $3$ modalities, $1,607$ subjects, and 2D plus 3D attack types. It is the first public dataset designed for exploring the impact of cross-ethnicity in the study of face anti-spoofing. As shown in Figure~\ref{fig:cefa}, samples from different ethnicities and several presentation attack types are included in the CASIA-SURF CeFA dataset.
Specifically, $4$ protocols are introduced to measure the impact under various evaluation conditions: cross-ethnicity (Protocol 1), (2) cross-PAI (Protocol 2), (3) cross-modality (Protocol 3) and (4) cross-ethnicity\&PAI (Protocol 4 as shown in Table~\ref{tab:cefa}). This paper is focus on most difficult one, Protocol 4. 

\subsection{Baseline Methods}

\noindent \textbf{SD-Net.}~CASIA provides a baseline for single-modal face anti-spoofing task via a ResNet-18-based~\cite{resnet} network named SD-Net~\cite{cefa}. It includes $3$ branches: static, dynamic and static-dynamic branches to learn hybrid features from static and dynamic images, respectively. For static and dynamic branches, each of them consists of $5$ res-blocks and $1$ GAP layer. 
A detailed description is provided in~\cite{cefa} for how to generate dynamic images. In short, a dynamic image is computed online with rank pooling based on $K$ consecutive frames. The reason for selection of dynamic images for rank pooling in SD-Net is that dynamic images have proved to have advantage over conventional optical flow~\cite{wang2017ordered,fernando2017rank}.%

\noindent \textbf{PSMM-Net.}~To make full use of multi-modal image data to alleviate the ethnic and attack bias, CASIA proposes a novel multi-modal fusion network, namely PSMM-Net~\cite{cefa}, which includes two main parts: a) the modality-specific network, contains three SD-Nets to respectively learn deep features from three modalities such as RGB, Depth, IR; b) a shared branch for all modalities, for the purpose of learning the complementary features among different modalities. They also design information exchange and interaction among SD-Nets and the shared branch, with the purpose of capturing correlations and complementary semantic information among different modalities.Two main kinds of losses are adopted to guide the training of PSMM-Net. One kind of loss corresponds to the losses of the three SD-Nets, \ie RGB, Depth and IR modalities data, denoted as $\mathcal L^{color}$, $\mathcal L^{depth}$ and $\mathcal L^{ir}$, respectively. Another kind of loss bases on the summed features from all SD-Nets and the shared branch, which guides the entire network training, denoted as $\mathcal L^{whole}$, which . The overall loss $\mathcal L$ of PSMM-Net is denoted as follow:

\begin{equation}\label{Eq:multi_modality_loss}
\mathcal{L} =  \mathcal{L}^{whole} + \mathcal{L}^{color} +  \mathcal{L}^{depth} + \mathcal{L}^{ir}
\end{equation}
\section{Methodology}
\label{method}
We focus on fusion network for CASIA-SURF CeFA dataset, since fusion network achieved better stability and robustness when facing various kinds of presentation attacks with previous CASIA-SURF dataset.
\subsection{The overall Model Architecture}
In this work, we propose a novel Pipeline-based CNN fusion architecture for multi-modal face anti-spoofing, named PipeNet. It is based on modified SENet154, with a Selective Modal Pipeline (SMP) module for multiple modalities of image input and a Limited Frame Vote (LFV) module for sequence input of video frames. 

Figure~\ref{fig:arch} shows the overall network architecture. The inputs are facial videos in $3$ modalities, which are RGB, depth and IR. For each modal, we take one frame as input and randomly crop it into patches, then send them to the corresponding modal pipeline. In each modal pipeline, data augmentation and CNN feature extraction are performed. The outputs of three pipelines are concatenated into one and sent to the fusion module for further feature abstraction. After linear connection, we obtain the predictions for each frame, and send all of them to the Limited Frame Vote module for iterative calculation. The output is the real face prediction probability of the original facial video.  

\subsection{Closing the Gap in Cross-ethnicity Learning}
As shown in Table~\ref{tab:cefa}, in the dataset CASIA SURF CeFA Protocal-4 (including 4@1,4@2,4@3), the faces are from different ethnicities between train and test set, which is called Cross-ethnicity learning. Figure~\ref{fig:cefa} shows the cropped faces of different ethnicities(African, East Asian and Central Asian) in $3$ modalities(RGB, depth and IR) in the case of real face. In intuitive observation, the biggest gap between ethnicities is skin tone in RGB modal. There is a potential risk of increasing test loss and ACER error rate, because test data is inconsistent with training data. To eliminate or reduce the gap in the input feature distributions between train and test set, we convert the RGB images of both set into other representations such as HSV, YCbCr and Grayscale, and pick the representation with best performance. The objective is to reduce the influence of skin tone information.   

\subsection{Selective Modal Pipeline for Different Modalities Data}
In essence, face anti-spoofing attack detection is an image classification task. It mainly contains two phases: CNN feature extraction with the first half CNN layers and feature map (binary) classification with other CNN layers. 

\begin{table}[htbp]
\begin{center}
\centering
\begin{tabular}{|l|c|c|c|}
\hline
Modalities & Crop face & Crop patches & Color trans \\
\hline\hline
 RGB & $\surd$ & $\surd$ & RGB2Gray\\
 \hline
 Depth & $\surd$ & $\surd$ & $\times$ \\
 \hline
 IR & $\surd$ & $\surd$ & $\times$\\
\hline
\end{tabular}
\end{center}
\caption{Selected augmentation methods for $3$ modalities.}
\label{tab:aug}
\end{table}

Because the face image in different modalities contains specific feature distribution and attributes, we treat them differently with selective pipeline in feature extraction phase. The \textit{pipeline} concept means it is an end-to-end data flow and it can be created and configured in a very flexible way; \textit{selective} means the structure of each pipeline is adapted to each modal. Pipelines for different modalities are independent which could be unified or specific.

We pick up basic block candidates from ResNet\cite{resnet}, ResNeXt\cite{resnext}, XceptionNet\cite{xception}, SENet\cite{se} \etal. and we align the input and output dimensions of these blocks to link them as a pipeline and deploy these pipelines to feature extraction module with input data from different modalities. We search for the best combination of three pipelines with experiments. Then, we concatenate the output from SMP as the input of the fusion module. The selective pipeline structure greatly improves the accuracy and efficiency in finding the most suitable structure. 

\subsection{Limited Frame Vote for Video Classification}
Since the input is in video-clip format, it is a sequence of continuous frames with different lengths. Accordingly, the proposed method adds a Limited Frame Vote (LFV) module to obtain the final statistical prediction of a video-clip input.

Several algorithm can provide a measurement of the central tendency of a probability distribution or the random variable characterized by that distribution. The performance mainly depends on the data distribution.
Mathematical expectation and geometric median are the most common methods for samples statistics.
\begin{figure}[htbp]
	\centering
	\includegraphics[height=3.5cm,width=8cm]{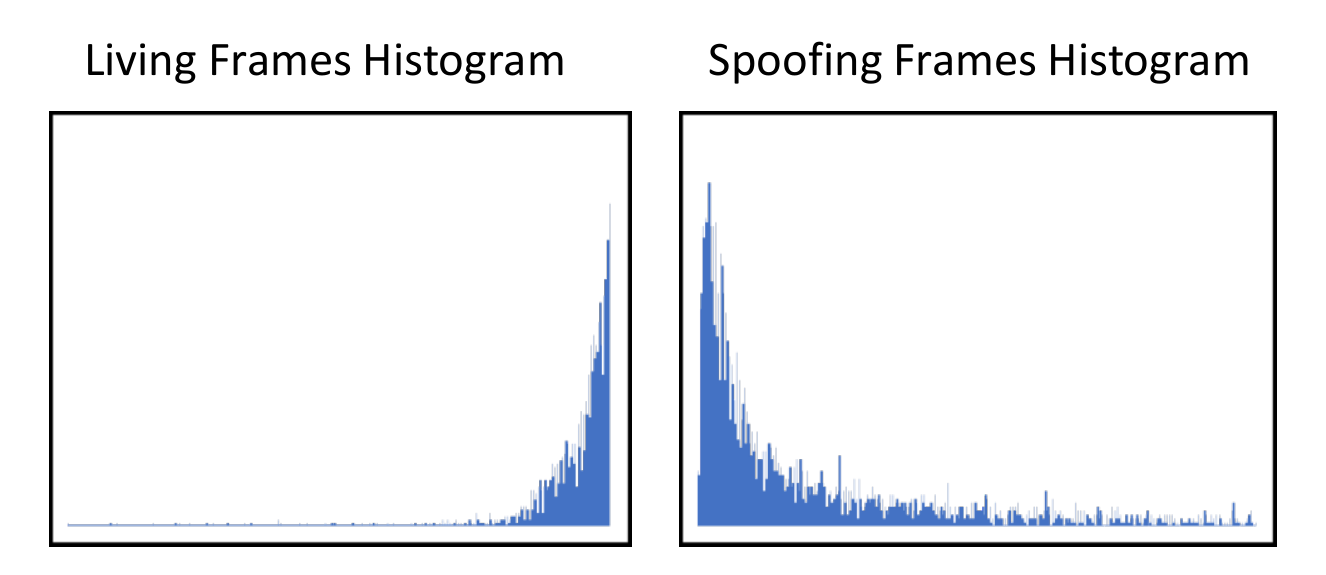}
	\caption{Frame Probability Distribution. }
\label{fig:hist}
\end{figure}


To avoid outliers, voting should be applied by the partial probabilities of frames. We create the following algorithm based on PauTa Criterion algorithm. Algorithm pseudocode is as Algorithm \ref{alg:3sigma}:

\begin{algorithm}[htbp]  
  \caption{Limited Frame Vote algorithm}  
  \label{alg:3sigma}  
  \begin{algorithmic}[1]  
    \Require $P$: list of frame prediction probabilities;
      
      $N$: frames length; 
      
      $\lambda$: $\sigma$ offset factor;
      
      $\tau$: $\sigma$ offset threshold;
    \Ensure  
      $E$
    \State initial $N$=length of whole frames in original clip
    \Repeat  
      \State compute expectation $E=\frac{\sum\limits_{i=1}^{N}P_{i}}{N}$;  
      \State compute std $\sigma=\sqrt{\frac{\sum\limits_{i=1}^{N}(P_{i}-E)^{2}}{N}}$;  
      \State filter $P$ with below equation and create new set to assign to $P$:
      $$
      P_{i} = 
        \begin{cases}
            P_{i} & P_{i} \in (E-\lambda*\sigma,E+\lambda*\sigma) \\
            0 & else
        \end{cases}
      $$
      \State compute number of Non-zero elements in $P$ and assign to $N$;  
    \Until  $\sigma \leq \tau $
  \end{algorithmic}  
\end{algorithm}  

In this algorithm, $P_{i}$ is input, $\lambda$ and $\tau$ are hard value from experience. 

\subsection{Some Other Tricks in Training}
There are also some training tricks doing benefit to our final accuracy. \ref{cd} is borrowed from winners~\cite{facebagnet,feathernets} in Face Anti-spoofing Attack Detection Challenge@CVPR2019 ~\cite{2019w}; ~\ref{cos} and ~\ref{loss} are breakthrough of previous classic works.

\subsubsection{Random Input Crop and Modal Dropout}
\label{cd}
During training, we apply patch-level input into Face Anti-spoofing area and prove that patch-level inputs can better extract spoof-specific discriminative information over the whole face area. And random dropout of one selective modal is adopted since it can avoid over-fitting and better make use of the characteristics in all three modalities. In the inference phase, we use fixed patches and full modalities to facilitate reproducing result.

\subsubsection{Dynamic Start Point for Cosine Decay Restarts}
\label{cos}
Cyclical cosine annealing learning rate~\cite{sgdr} is able to avoid falling into local minimum extreme point during gradient descent process. We improve it by using dynamic learning rate (LR) starting point for each cycle, instead of the fixed value in original version.

\begin{table*}[htbp]
\begin{center}
\centering
\begin{tabular}{|l|c|c|c|c|c|c|}
\hline
Leader &Team & Affiliation& APCER & BPCER & ACER & Rank \\
\hline\hline
Zitong Yu\cite{fisher1,fisher2} & BOBO & Oulu  &$1.05\pm 0.62$&$1.00\pm 0.66$ &$1.02\pm 0.59$&1\\
\hline
Zhihua Huang & Super & USTC &$0.62\pm 0.43$&$2.75\pm 1.50$&$1.68\pm 0.54$ &2\\
\hline
Qing Yang & Hulking(our team) & Intel &$3.25\pm 1.98$&$1.16\pm 1.12$&$2.21\pm 1.26$&3\\
\hline
\end{tabular}
\end{center}
\caption{Top-3 reproduced test set results by organizer in the final stage of Chalearn Multi-modal Cross-ethnicity Face anti-spoofing Recognition Challenge@CVPR2020. The mean and standard deviation values are calculated among results of \textit{Protocol 4}, 4@1, 4@2 and 4@3.}
\label{tab:benchmark}
\end{table*}


\subsubsection{Train/Validation Loss Balance Strategy}
\label{loss}
Snapshot ensemble~\cite{snapshot} inspired us on how to search for best snapshot point of the weight checkpoints during training process. We make a checkpoint-auto-save system whose input are ACER and train/validation loss. It automatically saves checkpoints in multiple positions in each cycle for potential global best. 

\section{Competition Details}
\label{competition}
\subsection{Dataset and Protocol}
We use CASIA-SURF CeFA dataset, which is introduced in Section~\ref{cefa}. In order to promote the competition and increase the level of challenge, the most challenging protocol \textit{Protocol 4} designed via combining the condition of both Protocol $1$ and $2$, was adopted as the evaluation criterion of the competition. As shown in Table~\ref{tab:cefa}, it contains $3$ data subsets: training, validation and testing sets containing $200$, $100$, and $200$ subjects, respectively. In addition, it has $3$ sub-protocols (\ie, $4@1$, $4@2$ and $4@3$) and in which one ethnicity is used for training and validation, and the remaining two ethnicities are used for testing. Moreover, the factor of PAIs are also considered in this protocol by setting different attack types during the training and testing phases.

\subsection{Evaluation Metrics}
From the evaluation of prediction results, we can get True Positive(TP), False Positive(FP), True Negative(TN), False Negative(FN).
Based on these four values, we can calculate the following evaluation metrics.
Donate Attack Presentation Classification Error Rate as:
\begin{equation}
APCER  =  FP / (TN + FP)
\label{eq:ap}
\end{equation}
Donate Bona Fide Presentation Classification Error Rate:
\begin{equation}
BPCER = FN / (FN + TP)
\label{eq:bpcer}
\end{equation}
Average Classification Error Rate is donated as:
\begin{equation}
ACER = (APCER + BPCER) / 2
\label{eq:acer}
\end{equation}

Eventually, the competition uses the ACER to determine the final ranking.

\subsection{Training Configuration}
We divided available data into train/validation set with 15:1 ratio with more detailed description in Section~\ref{ratio}. The optimization method is SGDR and learning rate is set to decay from $0.1$ to $0.001$. We train the model with $5$ cycles, each of which contains $100$ epochs. 

\subsection{Competition Results}
We won the 3rd place in the final ranking based on competition organizer's reproduced results. The final result of team scores and ranking is shown as Table~\ref{tab:benchmark}. In the Chalearn Multi-modal Cross-ethnicity Face Anti-spoofing Recognition Challenge@CVPR2020~\cite{2020w}, our final submission gets the score of $1.97\pm1.10$ ACER on the test set, while reproduced result is $2.21\pm1.26$ ACER, very close to our submission.
\begin{table}[htbp]
\begin{center}
\centering
\begin{tabular}{|l|c|c|c|}
\hline
Protocal & APCER & BPCER & ACER \\
\hline\hline
 4@1 & 1.44 & 0.00 & 0.72\\
 \hline
 4@2 & 4.11 & 1.50 & 2.80\\
 \hline
 4@3 & 2.55 & 2.24 & 2.40\\
\hline
\end{tabular}
\end{center}
\caption{Best Results of Three Models Corresponding to Protocol 4@1, 4@2 and 4@3.}
\label{tab:4_123}
\end{table}

Table~\ref{tab:4_123} shows the best scores of our submission for $3$ trained models corresponding to 4@1,4@2 and 4@3 protocols, which are $0.7222$, $2.8055$ and $2.4027$, respectively.

PipeNet results in Table~\ref{tab:benchmark} (Hulking) and Table~\ref{tab:4_123} are generated with same configurations.

\section{Experiments and Results Analysis}
\label{experiment}

\label{result}
\subsection{Effect of Different Train/Validation Segmentation Ratio}\label{ratio}
How to segment the training and validation dataset also affects the final result. Several ratios have been tried, including 15:1,12:1,10:1 and 3:1. We ultimately choose 15:1 which generates best performance among them. The test set is much larger than training and validation set and includes many non-existing presentation attacks types in training set. It is observed that a larger training set may still improve the model to learn more information about the unseen samples in test set. 

\subsection{Effect of Selective Modal Pipeline}
As shown in Table~\ref{tab:smp}, we have $5$ proposals for pipeline portfolio, including $4$ unified designs and $1$ selective design. \textit{Unified} means pipelines for RGB, depth and IR are exactly the same while \textit{selected} means different. We examine their ACER on dev set and test set simultaneously. In particular, we adjust the augmentation parameters and CNN blocks combination for each of the three pipelines. CNN blocks consist of SE-ResNetBottleneck and $3$ types of SE-ResNeXtBottleneck. The result shows that ACER is reduced after treating each model independently and customizing each modality with different pipeline.

For this section, we repeat each quantitative experiment three times for each of the three dataset protocols. The experiments include training and inference phases. We use the average value as the final result to ensure stability and consistency of model performance.
\begin{table}[htbp]
\begin{center}
\centering
\begin{tabular}{|l|c|c|c|c|}
\hline
\multirow{2}{*}{Pf.}&\multicolumn{3}{|c|}{Pipelines}&\multirow{2}{*}{ACER}\\
\cline{2-4}
 ~& RGB & Depth & IR &  \\
\hline\hline
 1 & SRB & SRB & SRB &$5.76\pm1.72$\\
 \hline
 2 & SRXB22 & SRXB22 & SRXB22 &$4.17\pm2.09$\\
 \hline
 3 & SRXB24 & SRXB24 & SRXB24 &$5.31\pm0.62$\\
 \hline
 4 & SRXB34 & SRXB34 & SRXB34 &$10.34\pm9.28$\\
 \hline
 5* & SRXB22 & SRB & SRXB22 &$1.97\pm1.10$\\
\hline
\end{tabular}
\end{center}
\caption{Results of Different Pipelines Fusion Portfolios. Portfolio 1-4 contain the same pipeline for RGB, depth and IR modalities, while Portfolio 5 customizes each modality with different pipeline. \textit{SRB} is short for SE-ResNetBottleneck; \textit{SRXB34} is short for SE-ResNeXtBottleneck, layer1 and layer2 repeat $3$ times and $4$ times (result of portfolio 5 is final submission score, better than the reproduced one in Table~\ref{tab:benchmark}).}
\label{tab:smp}
\end{table}

\subsection{Does Frame Time Sequence Order Matter?}
As shown in Table~\ref{tab:order}, before the decision of adopting LFV module, we perform experiments to measure the effect of sequence order on video classification results.
We take part of continuous frames from sample video as our input, which is corresponding to micro face motions or static face frames, from both living and spoofing video samples, respectively. We then modified PipeNet data augmentation module to process continuous frames. We rearrange the order of frame sequence and test it with three different orders - original order, reverse order and random order. N/A are the cases where dataset does not contain corresponding examples.
\begin{table}[htbp]
\begin{center}
\centering
\begin{tabular}{|l|c|c|c|c|}
\hline
\multirow{2}{*}{Protocol}&\multirow{2}{*}{Order}& \multicolumn{3}{|c|}{Probabilities} \\
\cline{3-5}
 & & Static & Smile & Blink \\
\hline\hline
 ~ &Original & 0.9999 & 0.9997 & 0.9999\\
 \cline{2-5} 
 Living &Reverse & 0.9999 & 0.9998 & 0.9999\\
 \cline{2-5}  
 ~ &Disorder  & 0.9999 & 0.9999 & 0.9999\\
 \hline
 ~ &Original &5.22e-6 & N/A & N/A\\
 \cline{2-5} 
 Print Photo&Reverse & 3.67e-6 & N/A & N/A\\
 \cline{2-5} 
 ~ &Disorder &3.00e-6 & N/A & N/A\\
 \hline
 ~ &Original &6.74e-5 & N/A & 0.0005\\
 \cline{2-5} 
 Video Replay &Reverse &5.06e-5 & N/A & 0.0002\\
 \cline{2-5} 
 ~ &Disorder& 5.64e-5 & N/A & 0.0002\\
\hline
 \multirow{3}{1.8cm}{3D Print Mask} &Original &0.0968 & N/A & N/A\\
 \cline{2-5} 
  &Reverse &0.1058 & N/A & N/A\\
 \cline{2-5} 
 ~ &Disorder& 0.0935 & N/A & N/A\\
\hline
\end{tabular}
\end{center}
\caption{Experiments on Effect of Input Frames' Order. N/A are the cases where dataset does not contain corresponding examples.}
\label{tab:order}
\end{table}

The probabilities of predictions among different order options are very close. Two possible reasons are: a) the order of frame sequence is not a key contribution factor in fusion network compared to RGB single-modal network; b) Even after basic cleaning and facial data alignment, the dataset frames are still not consistent and have various artifacts such as random transportation and partially cropped from background disturbs the context information.

As a result, we choose Limited Frames Vote strategy to obtain the probability of real face for each video sample.

\section{Conclusion}
\label{conclusion}
In this paper, we propose a flexible and practical multi-stream network architecture to build a robust face anti-spoofing system. The model, named PipeNet, is a pipeline-based design with Selective Modal Pipeline module and Limited Frame Vote module. The quantitative experiments show that the customized pipeline for each modality in PipeNet can make better use of different modalities data. In addition, LFV module provides stable and accurate prediction with continuous frames input.   
We apply the proposed PipeNet to Chalearn Multi-modal Cross-ethnicity Face Anti-spoofing Recognition Challenge@CVPR2020~\cite{2020w} and win the third place. Our final submission achieves a score of $2.21\pm1.26$ ACER on the test set. Our future research will focus on enhancing the self-adjustment capability of modal pipelines.



{\small
\bibliographystyle{ieee_fullname}
\bibliography{PipeNet}
}

\end{document}